\documentclass[runningheads]{llncs}
\usepackage[T1]{fontenc}
\usepackage{cite}
\usepackage{graphicx,verbatim}
\usepackage[colorlinks=true,
            linkcolor=red,
            urlcolor=red,
            citecolor=red]{hyperref}
\usepackage{amsmath}
\usepackage{amssymb}
\usepackage{bm}
\usepackage{booktabs}
\usepackage{multirow}
\usepackage[table]{xcolor}
\usepackage{colortbl}
\usepackage{threeparttable}
\usepackage{siunitx}
\usepackage{makecell}
\usepackage[misc]{ifsym}

\definecolor{engineerblue}{RGB}{221,235,247}
\newcommand{\ours}[1]{\cellcolor{engineerblue}\textbf{#1}}

\begin{document}
\title{Recurrent Contrastive Learning for \\Imbalanced Medical Image Classification}
\titlerunning{Recurrent Contrastive Learning}

\author{
Zhiyuan Zhu\inst{1,2}\thanks{Zhiyuan Zhu, Xinling Meng, Junxuan Yu and Jiongquan Chen contribute equally to this work. Corresponding emails: \email{luyongp@163.com} and \email{xinyang@szu.edu.cn}} \and 
Xinling Meng\inst{3\star} \and
Junxuan Yu\inst{1,2\star} \and
Jiongquan Chen\inst{1,2\star} \and
Qiongying Ni\inst{1,2} \and
Tuhang Shao\inst{1,2} \and
Yuhao Huang\inst{4} \and
Luping Zhou\inst{1,2} \and
Ruiyang Huang\inst{1,2} \and
Yuxue Wang\inst{5} \and
Rongliang Zhang\inst{5} \and
Xue Wang\inst{5} \and
Tianhong Tang\inst{6} \and
Likun Wang\inst{7} \and
Junbo Chen\inst{3} \and
Yong Jiang\inst{6}\and
Yongping Lu\inst{5}\textsuperscript{(\Letter)} \and
Xin Yang\inst{1,2}\textsuperscript{(\Letter)}
}

\authorrunning{Z. Zhu et al.}

\institute{
School of Biomedical Engineering, Medical School, Shenzhen University, China\\ 
\and
Ultra-X AI Lab, Shenzhen University, Guangdong, China
\and
School of Biomedical Engineering, South-Central Minzu University, Wuhan, China\\
\and
Department of Radiology, Boston Children's Hospital and Harvard Medical School, Boston, MA, USA\\ 
\and
The Affiliated Hospital of Yunnan University, Kunming, China\\
\and
Fuwai Shenzhen Hospital, Chinese Academy of Medical Sciences, Shenzhen, China\\
\and
First Affiliated Hospital of Hebei North University, Zhangjiakou, China\\
}

\maketitle
\begin{abstract}
Medical image classification often suffers from class imbalance due to the inherent disparities in disease incidence. Existing approaches, such as class resampling and loss reweighting, mainly improve learning within the observed feature distribution, but do not explicitly enlarge the latent support region of tail classes. As a result, tail-class representations remain overly compact and are easily encroached upon by head classes, leading to biased decision boundaries.
In this work, we propose \textbf{Recurrent Contrastive Learning (RCL)} for imbalanced medical image classification. RCL progressively expands the support region of tail classes by recurrently reusing historical feature states across training phases. Specifically, we adopt DINOv3 with LoRA adapters as the backbone to provide robust feature embeddings. We then devise a \textbf{Temporal Memory Queue (TMQ)} to preserve corpus-level features across training phases and provide diversified global references for contrastive learning. Based on TMQ, we construct \textbf{Temporal Anchors (TARs)} to form an anchor field around tail classes. This field enlarges the support region of tail classes, suppresses head-class encroachment, and improves inter-class separation. Extensive experiments on three imbalanced medical datasets demonstrate that RCL achieves consistent improvements over strong baselines. The code is available at \href{https://github.com/dndins/RCL}{https://github.com/dndins/RCL}.

\end{abstract}

\section{Introduction}
\begin{figure}
\includegraphics[width=\textwidth]{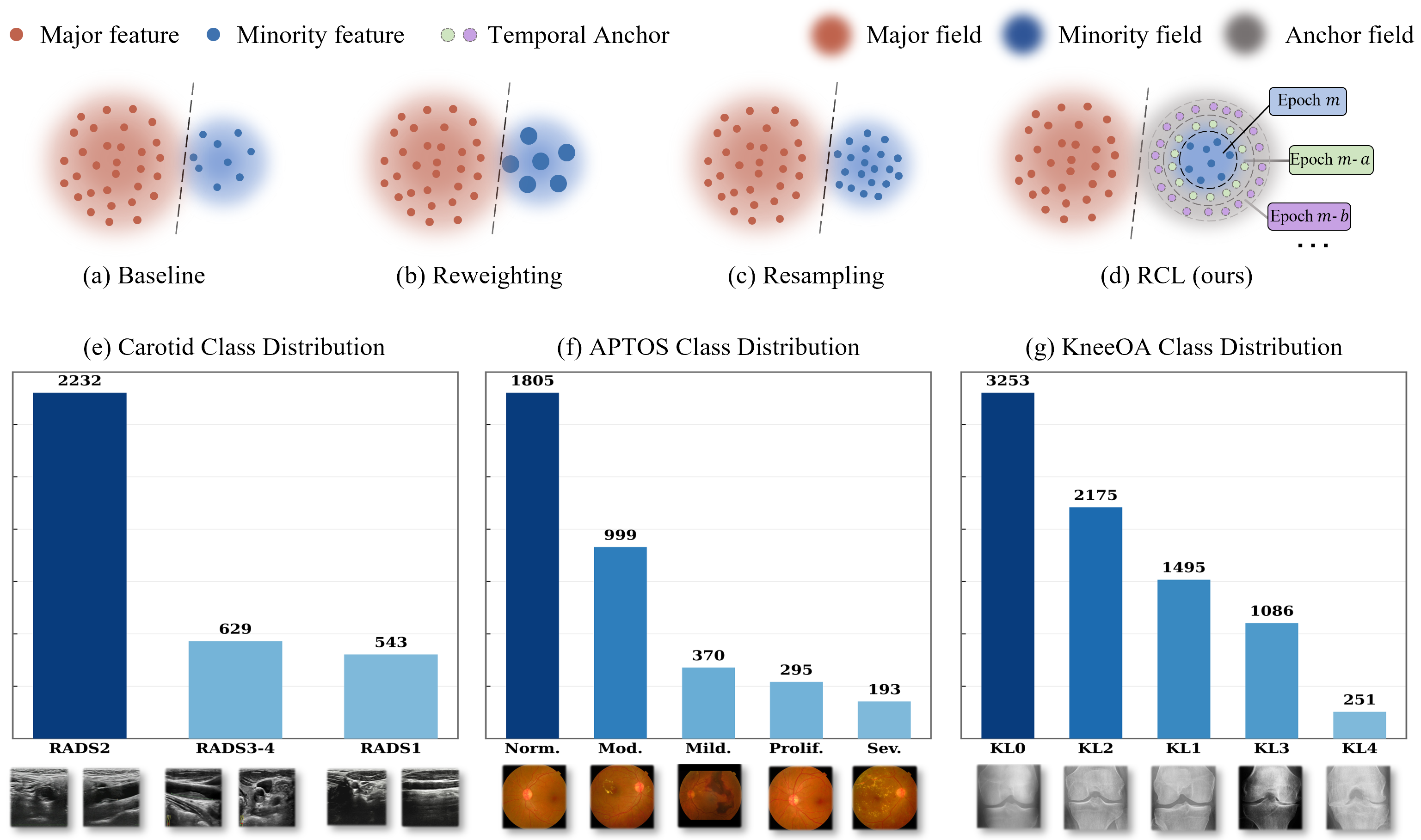}
\caption{Comparison of different imbalanced classification strategies and data distributions. (a): Data imbalance situation; (b-c): Data reweighting and resampling; (d): our anchor-based RCL; (e-g): The class distribution of three datasets} \label{fig1}
\end{figure}

Deep learning often benefits from large and balanced datasets to effectively learn and model feature distributions across different categories~\cite{imbalance-class}. However, in real-world medical scenarios, as shown in Fig. \ref{fig1} (a), the prevalence of different diseases varies substantially, resulting in a limited number of samples for rare conditions~\cite{rare-disease}. This imbalance biases model training toward head classes and weakens representation learning for tail classes~\cite{imbalance-learning}. Notably, tail classes often correspond to severe or rare disease types, where accurate identification is more critical for real-world clinical decision-making.

To address this class-imbalanced challenge, various strategies have been proposed in recent years. Menon et al.~\cite{LogitAdjustment} introduced a logit adjustment strategy that incorporates class-prior information into logits to correct prediction bias under class-imbalanced distributions, providing a theoretically grounded reweighting approach. 
To improve representation fairness in contrastive learning, Zhu et al.~\cite{BCL} reformulated supervised contrastive learning by balancing gradient contributions across classes to restore a more regular geometric structure in feature space. 
Beyond loss-level adjustments, Zhao et al.~\cite{neighbors} enhanced minority learning by incorporating neighboring tail categories to increase granularity. 
Wang et al.~\cite{MoE} proposed a mixture-of-experts framework to decouple category-specific knowledge into dedicated experts. 
In addition, Shao et al.~\cite{ADSR} designed a dual-axis style-based recalibration network to enhance minority feature discrimination through statistical loss constraints, while Zhou et al.~\cite{HiFuse} adopted hierarchical CNN–Transformer fusion to strengthen global–local feature modeling. 
For ordinal medical grading tasks, Tang et al.~\cite{DGN} modeled asymmetric Gaussian label distributions to better capture intra-class severity variations. 
Although these methods improve minority recognition from different perspectives, they primarily operate within the existing feature distribution through reweighting, resampling, or architectural modification (Fig.~\ref{fig1}(b, c)). They improve how features are used, but do not explicitly enlarge the support region of tail classes. Consequently, tail-class representations remain spatially compact and vulnerable to head-class encroachment, leading to reduced class separability.

In this study, we propose a novel learning framework named \textbf{Recurrent Contrastive Learning (RCL)}, which interprets class-imbalanced representation learning from a field-expansion perspective. As shown in Fig.~\ref{fig1}(d), we introduce \textbf{Temporal Anchors (TARs)}, which are recurrently derived from historical epoch features and used to occupy the surrounding region of tail-class features in latent space. These anchors expand the support region of tail classes and create a buffer margin against nearby head classes. To realize this mechanism, we maintain a \textbf{Temporal Memory Queue (TMQ)} to preserve corpus-level features from previous epochs and reuse them for anchor construction. By integrating TARs into contrastive learning, RCL progressively reshapes the feature geometry toward more balanced inter-class separation. Our contributions are threefold. \textit{First}, we propose RCL, an imbalanced medical image classification framework that enlarges tail-class support regions by injecting recurrently constructed temporal anchors into contrastive learning, mitigating head-class encroachment. \textit{Second}, we introduce TMQ to preserve corpus-level historical embeddings across training phases and use them to construct TARs for tail-class field expansion. \textit{Third}, as shown in Fig.~\ref{fig1}(e-g), we validate RCL on one multi-center private carotid dataset and two public datasets, KneeOA~\cite{OAI} and APTOS 2019~\cite{APTOS2019}, where it achieves consistent improvements over strong baselines. Experimental results demonstrate the effectiveness and generalization of the proposed method.

\section{Methodology}
\begin{figure}
\includegraphics[width=\textwidth]{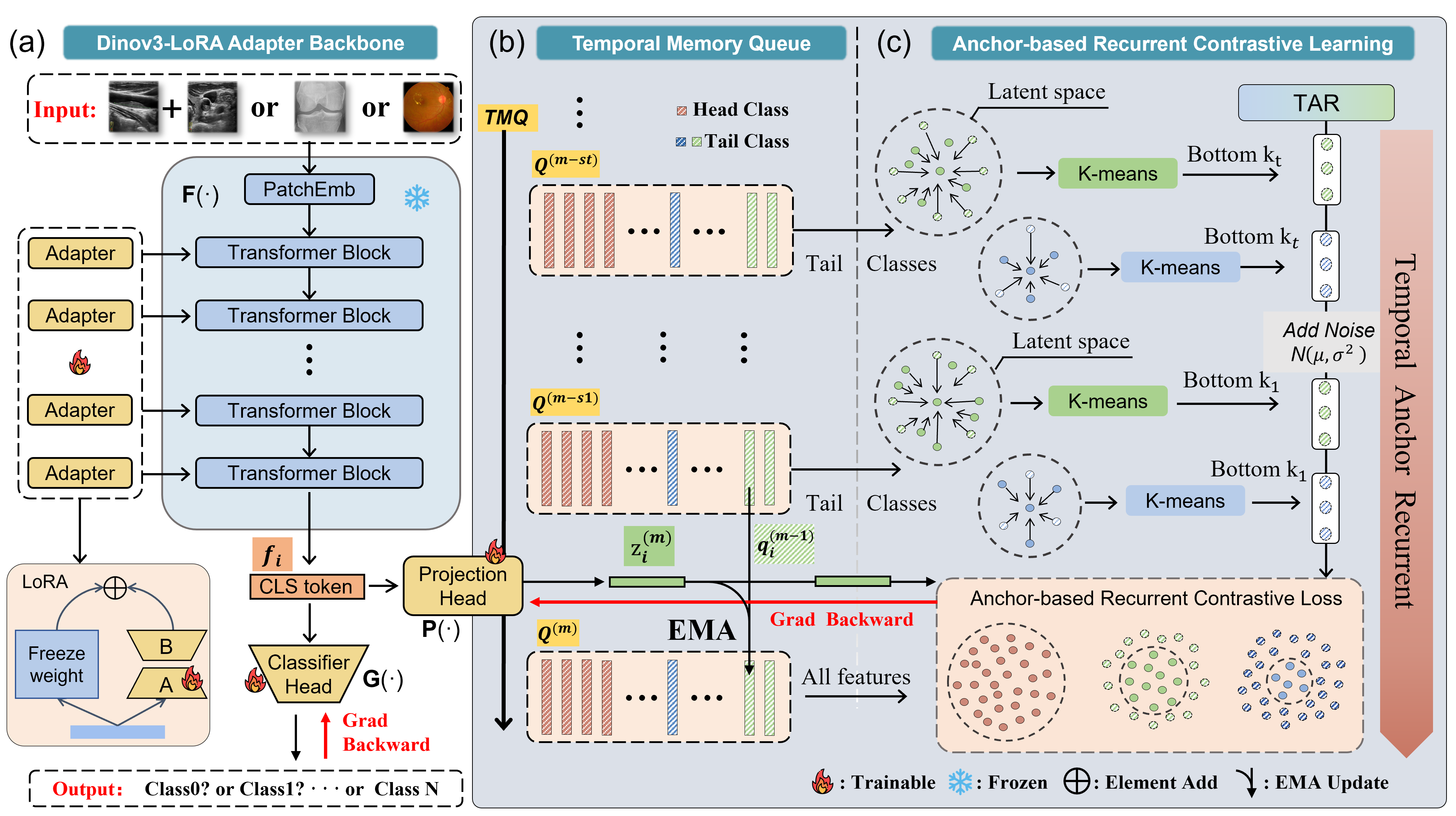}
\caption{Overview of the proposed RCL framework.} \label{fig2}
\end{figure}

Fig. ~\ref{fig2} provides an overview of RCL. For each input image \(x\), the DINOv3 backbone with LoRA adapters (rank=4) extracts a representation feature, which is used for both classification and projection into the latent space. RCL maintains a TMQ that stores features from previous training phases. We retrieve TARs from TMQ for tail classes and organize them into an \textbf{anchor field}. This field expands the latent support region of tail classes and serves as a buffer to enhance inter-class separability within RCL. More details are shown in the text below.

Formally, let $\mathcal{X} = \left \{ \left ( x_{i}, y_{i} \right ) | 1 \le  i \le  N\right \}$ be a class-imbalanced training set with \(C\) categories. The network consists of a backbone $\mathbf{F(\cdot )}$, a classifier head $\mathbf{G(\cdot )}$, and a projection head $\mathbf{P(\cdot )}$, and maintains stage-wise memory queues $\mathcal{Q}^{(m)} = \{ (q_{i}^{(m)}, y_{i})| 1\le i \le N \}, m = (0, 1, 2, ... M)$ as the anchor source.

\subsection{DINOv3 with LoRA Adapter Fine-tuning}
\label{sec:2.1}
The foundation model acquires strong representation capability through self-supervised learning on large-scale data, which enables effective adaptation to downstream domains with limited labeled data. As shown in Fig. \ref{fig2}(a), we feed an input image $x_{i}$ into backbone $\mathbf{F(\cdot )}$ and extract the final-layer 1280-dimensional class token $f_{i}$.  We employ a classifier head $\mathbf{G(\cdot )}$ that outputs predictions $p_i$ for cross-entropy supervision, and a projection head $\mathbf{P(\cdot )}$ that maps $f_i$ to a 256-dimensional normalized embedding $z_i$ for memory-bank update and contrastive learning. Both $\mathbf{G(\cdot )}$ and $\mathbf{P(\cdot )}$ adopt a Linear--ReLU--Linear design. During training, we freeze the DINOv3 backbone and optimize only the LoRA adapters with $\mathbf{G(\cdot )}$ and $\mathbf{P(\cdot )}$, enabling efficient adaptation with few trainable parameters.

\subsection{Temporal Memory Queue}
\label{sec:2.2}
Contrastive learning has shown clear benefits in imbalanced medical image classification tasks~\cite{cl-1, cl-3}. However, many existing methods perform contrastive matching only at the batch-level or within a FIFO queue, which limits access to global and temporally diverse feature references. Inspired by~\cite{bank-seg, CVCRF}, we introduce a TMQ that retains multiple memory banks across training stages and provides diversified global references for contrastive learning. Specifically, in epoch \(0\), we extract the projection embedding as \(z_{i}^{(0)}\) for each training sample \(x_{i}\), and use \(\{(z_{i}^{(0)}, y_{i})\mid 1\le i\le N\}\) to initialize the memory bank \(\mathcal{Q}^{(0)}\), which is defined as:
\begin{equation}
\mathcal{Q}^{(0)} = \{ (q_{i}^{(0)}, y_{i})\mid 1\le i \le N \} = \{ (z_{i}^{(0)}, y_{i})\mid 1\le i \le N \}.
\label{eq:init_mb}
\end{equation}
In the subsequent epoch \(m\), as shown in Fig. ~\ref{fig2}(b), the projection embedding \(z_{i}^{(m)}\) is used to update \(q_{i}^{(m)}\) in \(\mathcal{Q}^{(m)}\) via exponential moving average:
\begin{equation}
q_{i}^{(m)} = (1 - \lambda) q_{i}^{(m-1)} + \lambda z_{i}^{(m)}.
\label{eq:ema}
\end{equation}
where \(\lambda \in [0,1)\) is the EMA momentum controlling the contribution of the current embedding. Importantly, we retain the updated bank \(\mathcal{Q}^{(m)}\) after each epoch and enqueue it over epochs, forming the $TMQ = \{\mathcal{Q}^{(m)}\}_{m=0}^{M}$, which enables recurrent access to historical global embeddings.

This design provides two key advantages. First, each epoch-wise bank \(\mathcal{Q}^{(m)}\) approximates the global feature distribution at a specific optimization stage, and EMA updates keep stored features aligned with the evolving representation space, enabling richer and more reliable positives/negatives for contrastive learning. Second, by preserving multiple historical banks, TMQ serves as a carrier of temporal knowledge across training phases, which will be leveraged for temporal anchor-based field construction in Sec.~\ref{sec:2.3}.

\subsection{Anchor-based Recurrent Contrastive Learning}
\label{sec:2.3}
The temporal memory queue $TMQ = \{\mathcal{Q}^{(m)}\}_{m=0}^{M}$ is introduced to provide a global view of the feature distribution, enabling the model to learn dataset-level representations. However, the features stored in $Q^{(m)}$ still follow a class-imbalanced distribution. To alleviate this inter-class imbalance, we retrieve TARs from historical memory banks to enrich tail-class support and promote balanced learning.

As illustrated in Fig. \ref{fig2}(b, c), for a query $z_{i}^{(m)}$ with label $c$ at epoch $m$, we compute the contrastive loss $L_{RCL}$ using features in memory bank queue $Q$. 

\noindent \textbf{In current epoch $m$}, the positive set $Q_{+}^{(m)}$ consists of features sharing same class label as $z_{i}$, while the negative set $Q_{-}^{(m)}$ contains features from other classes:
\begin{equation}
Q_{+}^{(m)} = \{q_{i+}^{(m)}| y_{i} = c\}, \qquad
Q_{-}^{(m)} = \{q_{i-}^{(m)}| y_{i} \ne  c\}.
\label{eq:Qm}
\end{equation}
\noindent \textbf{In earlier epoch $m-st$}, we collect candidate tail-class features as $TAR^{(m-st)}$ from $Q^{(m-st)}$. Specifically, for each tail class in $Q^{(m-st)}$, we apply K-means selection to intra-class features, compute each feature’s similarity to the corresponding cluster center, and select the bottom-$k_{t}$ features. These selected features are then perturbed with Gaussian noise as in Eq.~\ref{eq:SAT_m-t}:
\begin{equation}
TAR^{(m-st)} = Bottom\text{-}K \big( k-means(\{(q_i^{(m-st)}, y_i) \mid y_i \in \mathcal{T}\}) \big) + \varepsilon \sim \mathcal{N}(0,\,0.1^2).
\label{eq:SAT_m-t}
\end{equation}
where $\mathcal{T}$ denotes the set of tail classes. Aggregating anchors from multiple previous epochs yields the final anchor set $A$:
\begin{equation}
A =  \bigcup_{t=1}^{n} TAR^{(m\text{-}st)} = \{(a_j, y_j)\mid y_j \in \mathcal{T}\}.
\label{eq:A_set}
\end{equation}
where $s$ is the sample step, $n$ is the sample number. Accordingly, the anchor-positive and anchor-negative sets are defined as:
\[
A_{+} = \{(a_{j+}, y_{j})| y_{j} = c \},  \qquad
A_{-} = \{(a_{j-}, y_{j})| y_{j} \neq c \}.
\]
The recurrent contrastive loss is defined as:
\begin{equation}
\mathcal{L}_{\text{RCL}}
=
-
\log
\frac{
\sum\limits_{q \in Q_+^{(m)}}
\exp\!\left(
\frac{\mathrm{sim}(z_i^{(m)}, q)}{\beta}
\right)
+
\sum\limits_{a \in A_+}
\exp\!\left(
\frac{\mathrm{sim}(z_i^{(m)}, a)}{\beta}
\right)
}{
\sum\limits_{q \in Q^{(m)}}
\exp\!\left(
\frac{\mathrm{sim}(z_i^{(m)}, q)}{\beta}
\right)
+
\sum\limits_{a \in A}
\exp\!\left(
\frac{\mathrm{sim}(z_i^{(m)}, a)}{\beta}
\right)
}.
\end{equation}
The final loss $L$ is defined as $L = L_{ce} + \alpha L_{RCL}$, where $\alpha$ denotes the loss balancing coefficient and \(\beta\) is the contrastive temperature.

\section{Experimental Results}
{\bfseries Datasets and Implementations.}
To evaluate RCL, we collected a private multi-center carotid ultrasound dataset with 3,404 paired images (543 RADS1, 2,232 RADS2, and 629 RADS3–4) from 12 centers, graded according to carotid plaque RADS guidelines~\cite{P-RADS}. The study was approved by the local Institutional Review Board (Anonymized ID: SP2023135 (01)). To verify generalizability, we further evaluated RCL on two public imbalanced datasets: APTOS 2019~\cite{APTOS2019} for diabetic retinopathy grading (3,662 images) and KneeOA~\cite{OAI} for Kellgren–Lawrence (KL) grading (8,260 images). All images were padded to a square shape and then resized to $256 \times 256$. The Carotid and APTOS 2019 datasets were split at an 8:2 ratio into training and test sets, with four-fold cross-validation performed on the training set and mean test performance reported. The split was conducted at the patient level to prevent data leakage. KneeOA followed the official split protocol. RCL was implemented in PyTorch and trained for 100 epochs on an NVIDIA GeForce RTX 3090 GPU using AdamW (initial learning rate $1 \times 10^{-4}$). The EMA momentum $\lambda$ and contrastive temperature $\beta$ were set to 0.01 and 0.1, respectively. The loss weight $\alpha$ was linearly decayed from 1 to 0.1. The sample step $s$ was set as 3. Among the $n$ sampled historical banks $\{Q^{(m-s)}, Q^{(m-2s)}, \ldots, Q^{(m-ns)}\}$, the bottom selection ratio $k_t$ was linearly decreased from 50\% for the most recent bank to 10\% for the earliest, so that temporally closer banks contribute more anchor candidates. Five standard augmentation operations from the $timm$ were randomly applied during training.
\begin{table*}[htbp]
\footnotesize
\centering
\setlength{\tabcolsep}{1.4pt}
\renewcommand{\arraystretch}{1.1}
\caption{Performance comparison and ablation on three datasets. 
BAcc(\%), F1(\%), Acc(\%), and QWK (three decimals). Best results (\textbf{bold}), second-best (\underline{underlined}).}
\label{tab:main_results}
\resizebox{\textwidth}{!}{
\begin{tabular}{@{}lcccccccccccc@{}}
\toprule
\multirow{2}{*}{Method} 
& \multicolumn{4}{c}{Carotid} 
& \multicolumn{4}{c}{APTOS 2019} 
& \multicolumn{4}{c}{KneeOA} \\
\cmidrule(lr){2-5} 
\cmidrule(lr){6-9} 
\cmidrule(lr){10-13}
& Bacc$\uparrow$ & F1$\uparrow$ & Acc$\uparrow$ & QWK$\uparrow$ 
& Bacc$\uparrow$ & F1$\uparrow$ & Acc$\uparrow$ & QWK$\uparrow$
& Bacc$\uparrow$ & F1$\uparrow$ & Acc$\uparrow$ & QWK$\uparrow$ \\
\midrule

ResNet50\cite{resnet}
& 74.97 & 75.73 & 80.65 & 0.681
& 50.66 & 50.91 & 72.50 & 0.700
& 59.13 & 58.95 & 58.03 & 0.729
\\

Focal$^{*}$\cite{focal}
& 76.01 & 77.62 & 82.55 & 0.731
& 52.08 & 52.38 & 73.46 & 0.711
& 60.16 & 62.47 & 61.35 & 0.744
\\

BCL$^{*}$\cite{BCL}
& 85.02 & 79.22 & 81.09 & 0.763
& 58.17 & 50.74 & 68.13 & 0.757
& \underline{69.58} & 67.98 & 66.55 & 0.802
\\

ECL\cite{ECL}
& 82.86 & 83.41 & 86.95 & 0.790
& 64.85 & 66.71 & 83.31 & 0.890
& 68.24 & \underline{68.81} & 67.19 & 0.787
\\

HiFuse\cite{HiFuse}
& 83.58 & 83.71 & 86.51 & 0.801
& 66.98 & 67.29 & 81.94 & 0.840
& 61.59 & 57.70 & 63.10 & 0.725
\\

ADSR\cite{ADSR}
& 84.60 & 82.65 & 85.52 & 0.786
& 62.56 & 64.35 & 81.53 & 0.855
& 65.21 & 64.51 & 64.31 & 0.778
\\

DGN \cite{DGN}
& 81.65 & 83.26 & 87.10 & 0.779
& 67.54 & 68.51 & 83.58 & 0.898
& 62.56 & 62.26 & \underline{68.84} & 0.813
\\

\midrule

DINOv3
& 79.75 & 82.33 & 86.14 & 0.793
& \underline{68.48} & \underline{69.20} & \underline{83.75} & \underline{0.920}
& 61.10 & 61.82 & 63.35 & 0.611
\\

RCL$^{*}$
& \underline{86.92} & \ours{88.10} & \ours{90.32} & \ours{0.853}
& 66.23 & 68.21 & 83.72 & 0.894
& 66.65 & 66.65 & 67.87 & \underline{0.825}
\\

DINOv3$+$RCL
& \ours{87.10} & \underline{87.87} & \underline{90.06} & \underline{0.852}
& \ours{69.01} & \ours{70.79} & \ours{85.39} & \ours{0.922}
& \ours{70.19} & \ours{70.36} & \ours{70.77} & \ours{0.843}
\\

\bottomrule
\end{tabular}
}

\begin{tablenotes}[flushleft]
\footnotesize
\scriptsize
\item * using ResNet50 as the backbone.
\end{tablenotes}

\end{table*}

\noindent{\bfseries Quantitative analysis.}
Balanced Accuracy (Bacc), F1-score (F1), Accuracy (Acc), and Quadratic Weighted Kappa (QWK) were adopted to comprehensively evaluate performance under class-imbalanced settings.
Table~\ref{tab:main_results} presents both quantitative comparisons with state-of-the-art class-imbalanced learning approaches ~\cite{resnet,  HiFuse, focal, BCL, ECL, ADSR, DGN} and the ablation study. Our framework achieves the best Bacc on all three datasets, reaching 87.10\%, 69.01\%, and 70.19\% on Carotid, APTOS 2019, and KneeOA, respectively, outperforming all competing methods. Compared with vanilla DINOv3 fine-tuning, DINOv3+RCL improves Bacc by 7.35\% on Carotid and 9.09\% on KneeOA, demonstrating consistent gains across different imbalance scenarios. On the APTOS 2019, the DINOv3-based RCL achieves a QWK of 0.92, surpassing the top-ranked CNN-Ensemble method ~\cite{APTOS2019} (QWK = 0.91) in the original challenge, while the ResNet50-based RCL also exceeds the second-ranked Inception-v4 model ~\cite{APTOS2019} (QWK = 0.88). When we replace DINOv3 with ResNet50 (the second-to-last row), RCL also yields substantial improvements. On the Carotid and KneeOA datasets, RCL based on ResNet50 even surpasses the performance of DINOv3 fine-tuning, indicating that RCL enhances feature discrimination beyond the backbone alone.

{\setlength{\textfloatsep}{10pt}        
 \setlength{\floatsep}{10pt}            
 \setlength{\intextsep}{10pt}           
 \setlength{\abovecaptionskip}{3pt}    
 \setlength{\belowcaptionskip}{-3pt}   

\begin{table}[!ht]
\centering
\scriptsize
\renewcommand{\arraystretch}{1.05}

\begin{threeparttable}
\caption{Effect of different sample numbers and tail-class settings on the Carotid dataset. Best results (\textbf{bold}), second-best (\underline{underlined}).}
\label{tab:minor_epoch_carotid}
\begin{tabular*}{\columnwidth}{@{\extracolsep{\fill}} c c cccc c cccc}
\toprule
$n$* &
Tail Classes &
Bacc$\uparrow$ & F1$\uparrow$ & Acc$\uparrow$ & QWK$\uparrow$ &
Tail Classes &
Bacc$\uparrow$ & F1$\uparrow$ & Acc$\uparrow$ & QWK$\uparrow$ \\
\midrule

3 & \multirow{3}{*}{1}
  & 86.86 & \underline{88.78} & \textbf{91.06} & \textbf{0.862}
  & \multirow{3}{*}{2}
  & \underline{88.57} & 88.68 & \underline{90.47} & 0.860 \\

5 &
  & 87.72 & 88.56 & \underline{90.47} & 0.858
  &
  & 87.42 & 87.70 & 89.59 & 0.847 \\

7 &
  & 86.85 & 87.34 & 89.59 & 0.839
  &
  & \textbf{88.68} & \textbf{89.18} & \textbf{91.06} & \underline{0.861} \\

\bottomrule
\end{tabular*}

\begin{tablenotes}[flushleft]
\footnotesize
\scriptsize
\item *$n$: Sample Number.
\end{tablenotes}

\end{threeparttable}
\end{table}

Table~\ref{tab:minor_epoch_carotid} reports the effect of different sample numbers and tail-class settings on the Carotid dataset. Owing to space constraints, we present the hyperparameter configurations that show the most significant influence on the experimental results. Here, Sample Number $n$ denotes the number of historical memory banks sampled to construct anchors. Tail Classes denotes the number of tail classes involved as TARs. The setting with Tail Classes = 2 and $n$ = 7 achieves the best overall performance, yielding the highest Bacc (88.68\%) and F1 (89.18\%), together with the best overall Acc (91.06\%). By contrast, the highest QWK is obtained with Tail Classes = 1 and $n$ = 3 (0.862). Overall, using two tail classes leads to stronger and more stable performance across different sample epochs, suggesting that richer tail-class information and progressively refined historical features are beneficial for robust representation learning under class imbalance.

\begin{figure}[t]
\includegraphics[width=\textwidth]{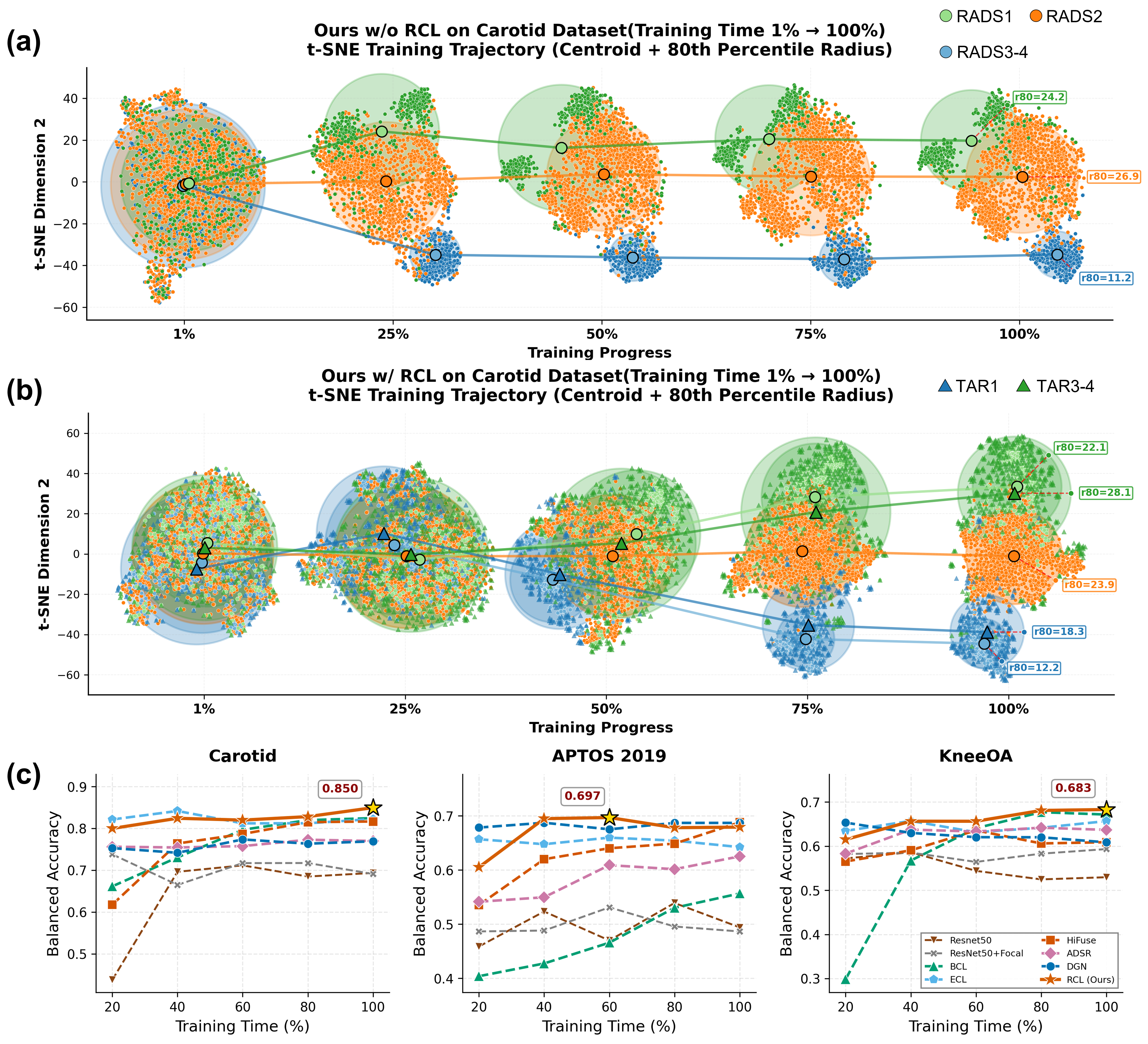}
\caption{t-SNE visualization and Bacc evolution at different training stages.} \label{fig:fig3}
\end{figure}

\textbf{Qualitative analysis.} We visualize t-SNE embeddings at multiple training phases in Fig.~\ref{fig:fig3}(a, b), summarizing each class by its centroid trajectory and 80th-percentile radius. Fig.~\ref{fig:fig3}(a) shows the DINOv3 training without RCL, whereas Fig.~\ref{fig:fig3}(b) shows with RCL enabled. Compared with the DINOv3-baseline, three phenomena emerge under RCL: 
(i) Compared with Fig.~\ref{fig:fig3}(a), the relative field of head and tail classes becomes more balanced in Fig.~\ref{fig:fig3}(b), indicating a fairer distribution of feature support in latent space; 
(ii) TARs display a larger radius than current epoch features, suggesting an expanded tail-class boundary induced by the anchor field;
and (iii) TAR1 field occupies the intermediate region between RADS2 and RADS1 fields, forming an effective buffer area. This buffer mechanism pushes RADS1 centers away from head-class distributions, thereby enlarging inter-class margins.
These observations are all consistent with the field-guided mechanism of RCL.
Fig.~\ref{fig:fig3}(c) reports validation Balanced Accuracy for checkpoints from different epochs. RCL achieves the best late-epoch performance on Carotid, APTOS 2019, and KneeOA, consistent with the progressively stabilized separation observed in the embedding dynamics.

\section{Conclusion}
In this work, we propose a novel Recurrent Contrastive Learning for imbalanced medical image classification. By leveraging a Temporal Memory Queue and constructing Temporal Anchors from historical feature states, RCL forms an anchor field to expand the support region of tail classes and mitigate head-class dominance. However, the framework relies on manual hyperparameter tuning. Future work will explore adaptive anchor selection to address this limitation.

\begin{credits}
\subsubsection{\ackname} This study was funded by the Yunnan Major Science and Technology Special Project Program (No. 202402AA310052), the Yunnan Key Research and Development Program(202503AP140014), and the Guangxi Province Science Program (No.2024AB17023).

\subsubsection{\discintname}
The authors have no competing interests to declare that are relevant to the content of this article.
\end{credits}

\bibliographystyle{plain}
\bibliography{references}

\end{document}